\documentclass[conference]{IEEEtran}
\IEEEoverridecommandlockouts
\usepackage{cite}
\usepackage{array}
\usepackage{amsmath,amssymb,amsfonts}
\usepackage{algorithmic}
\usepackage{dblfloatfix}
\usepackage{graphicx}
\usepackage{textcomp}
\usepackage{booktabs}
\usepackage{xcolor}
\usepackage{hyperref}
\usepackage{subcaption}
\usepackage{tikz}

\newcolumntype{P}[1]{>{\centering\arraybackslash}p{#1}}
\newcolumntype{M}[1]{>{\centering\arraybackslash}m{#1}}

\def\BibTeX{{\rm B\kern-.05em{\sc i\kern-.025em b}\kern-.08em
    T\kern-.1667em\lower.7ex\hbox{E}\kern-.125emX}}
    
\usepackage[absolute,showboxes]{textpos}

\TPMargin{5pt}

\newcommand{\copyrightstatement}{
    \begin{textblock}{0.80}(0.1,0.93)    
{~\copyright~2020 IEEE}
  Personal use of this material is permitted.  Permission from IEEE must be obtained for all other uses, in any current or future media, including reprinting/republishing this material for advertising or promotional purposes, creating new collective works, for resale or redistribution to servers or lists, or reuse of any copyrighted component of this work in other works.
    \end{textblock}
}    

\begin{document}
\copyrightstatement
\bstctlcite{IEEEexample:BSTcontrol}

\makeatletter
\def\ps@IEEEtitlepagestyle{
  \def\@oddfoot{\mycopyrightnotice}
  \def\@evenfoot{}
}

\newcommand\copyrighttext{%
  {~\copyright~2020 IEEE}
  Personal use of this material is permitted.  Permission from IEEE must be obtained for all other uses, in any current or future media, including reprinting/republishing this material for advertising or promotional purposes, creating new collective works, for resale or redistribution to servers or lists, or reuse of any copyrighted component of this work in other works.}

\def\mycopyrightnotice{
  \gdef\mycopyrightnotice{}
}

\title{Multi-Attributed and Structured Text-to-Face Synthesis\\
}

\author{\IEEEauthorblockN{1\textsuperscript{st} Rohan Wadhawan*} \IEEEauthorblockA{\textit{Department of Computer Science and Engineering} \\ \textit{Netaji Subhas Institute of Technology}\\ New Delhi, India \\ rohanwadhawan7@gmail.com} 

\and 

\IEEEauthorblockN{1\textsuperscript{st} Tanuj Drall*} \IEEEauthorblockA{\textit{Department of Computer Science and Engineering} \\ \textit{Netaji Subhas Institute of Technology}\\ New Delhi, India  \\ tanuj.drall@gmail.com}

\and 

\IEEEauthorblockN{2\textsuperscript{nd} Shubham Singh} \IEEEauthorblockA{\textit{Department of Computer Science and Engineering} \\ \textit{Netaji Subhas Institute of Technology}\\ New Delhi, India \\ vshubham5636@gmail.com} 
\and

\IEEEauthorblockN{3\textsuperscript{rd} Shampa Chakraverty} \IEEEauthorblockA{\textit{Department of Computer Science and Engineering} \\ \textit{Netaji Subhas University of Technology}\\ New Delhi, India  \\ shampa@nsut.ac.in}
}

\maketitle
\footnotetext{\textsuperscript{*}Equal Contribution}

\begin{abstract}
Generative Adversarial Networks (GANs) have revolutionized image synthesis through many applications like face generation, photograph editing, and image super-resolution. Image synthesis using GANs has predominantly been uni-modal, with few approaches that can synthesize images from text or other data modes. Text-to-image synthesis, especially text-to-face synthesis, has promising use cases of robust face-generation from eye witness accounts and augmentation of the reading experience with visual cues. However, only a couple of datasets provide consolidated face data and textual descriptions for text-to-face synthesis. Moreover, these textual annotations are less extensive and descriptive, which reduces the diversity of faces generated from it. This paper empirically proves that increasing the number of facial attributes in each textual description helps GANs generate more diverse and real-looking faces. To prove this, we propose a new methodology that focuses on using structured textual descriptions. We also consolidate a Multi-Attributed and Structured Text-to-face (MAST) dataset consisting of high-quality images with structured textual annotations and make it available to researchers to experiment and build upon. Lastly, we report benchmark Fr\'echet's Inception Distance (FID), Facial Semantic Similarity (FSS), and Facial Semantic Distance (FSD) scores for the MAST dataset.
\end{abstract}


\section{Introduction}
Generative Adversarial Networks (GAN) have been successfully employed for the artificial generation of images, especially faces \cite{goodfellow2014generative,radford2015unsupervised,karras2017progressive,heusel2017gans,karras2019style,karnewar2019msg}. GANs have been conditioned on image labels, metadata, or textual descriptions to exercise control over the generation process. \cite{mirza2014conditional,chen2016infogan,odena2017conditional}. A major motivation behind \textbf{Multi-Attributed and Structured Text-to-Face Synthesis} is to provide an unbiased and detailed facial generation of suspects from eye-witness accounts. In addition to GANs, the field of facial generation is being revolutionized by datasets like LFW \cite{huang2008labeled}, CelebA \cite{liu2015deep}, CelebA-HQ \cite{karras2017progressive}, Face2Text \cite{gatt2018face2text}, SCU-Text2Face \cite{chen2019ftgan}, and so on. Some of these datasets also provide information about facial features present in the images as binary attributes or textual annotations. A combination of binary attributes is suitable for generating detailed images. However, this approach does not capture the relationship among attributes, resulting in poor global structure images. Textual annotations represent these relationships well, but they suffer from a scarcity of attributes.

This paper empirically proves that increasing the number of facial attributes in each textual description helps GANs generate more diverse and real-looking faces. We develop a new methodology of structured textual descriptions that incorporates a large number of facial features in each description. This methodology combines the detailed information present in binary attributes with the global structure provided by textual annotation. In the \textbf{structured approach}, a face is annotated feature-wise to accumulate a large number of attributes in each annotation. On the other hand, the resultant annotations usually consist of only a few attributes due to the absence of feature-wise annotation and the dependence on the annotator's observational skill to label facial features in the unstructured approach. As the diversity of features learned by the GAN conditioned on a textual description increases with an increase in the number of attributes in the description, the structured approach promotes the learning of various facial features.

Further, we curate a face dataset of 1993 high-resolution face images with detailed structured textual descriptions and name it \textbf{Multi-Attributed and Structured Text-to-face (MAST)} dataset. \textbf{Each textual annotation consists of 15 facial attributes} on an average from a complete set of 30 attributes consolidated from our crowdsourcing platform, CelebA-HQ dataset, and Microsoft Face API. Since the aim is to develop a textual facial attribute to a visual feature mapping, we neglect redundant stop words like pronouns and helping verbs in the annotations. We apply the AttnGAN \cite{xu2018attngan} network to perform text-to-face synthesis using the MAST dataset. Furthermore, we measure the quality of the generated faces using FID \cite{heusel2017gans}, and the similarity with their corresponding ground truth using FSS and FSD \cite{chen2019ftgan}. Finally, we report the \textbf{benchmark scores of 54.09 FID, 1.080 FSD, and 60.42\% FSS} achieved by training the AttnGAN on the MAST dataset.

This paper makes the following contributions: (i) We develop a new methodology of structured textual descriptions that incorporates a large number of facial features in each description and curate the MAST dataset to show its effectiveness. (ii) We demonstrate that these descriptions help us generate fine-grained realistic-looking faces from the text and report the \textbf{benchmark scores of 54.09 FID, 1.080 FSD, and 60.42\% FSS} achieved by training the AttnGAN on the MAST dataset.

\begin{figure}[!t]
    \centering
    \includegraphics[width=\columnwidth, height=1.5cm]{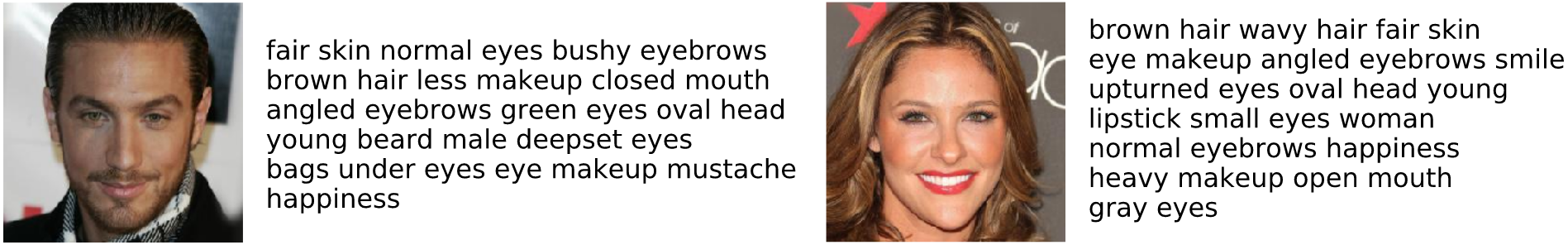}
    \caption{Ground truth face images from MAST dataset along with their textual descriptions}
    \label{fig:sample-dataset}
    \vspace{-0.4cm}
\end{figure}

\section{Related Work}
\paragraph{Text-to-Image synthesis} employs a text encoder and an image decoder framework to generate images. The use of Char-CNN-RNN and Word-CNN-RNN encoders to produce multi-modal context embeddings significantly enhanced the single-step text-to-image synthesis using GANs \cite{reed2016generative}. StackGAN transformed this single-stage generation process into a multi-stage process. The initial stages sketched primitive shapes and primary colors of the image, and the final stages converted the intermediate low-resolution image to a high-resolution photo-realistic image \cite{zhang2017stackgan, zhang2018stackgan++}. AttnGAN enhanced this multi-stage generation process through the use of the attention mechanism, which incorporated fine-grained word-level information of the text in addition to its global sentence-level information.

\paragraph{Text-to-face synthesis} A repository named T2F on Github(https://github.com/akanimax/T2F) employed a hybrid of ProGAN \cite{karras2017progressive}, and StackGAN \cite{zhang2017stackgan, zhang2018stackgan++} model to generate faces using the Face2Text dataset, which consists of 400 images taken from the LFW dataset with textual annotations consisting of \textbf{3-5 facial attributes per annotation}. Due to the small number of samples and scarcity of attributes in annotations, the face synthesized were of low quality. The SCU-Text2face dataset improved upon it by increasing the number of samples to 1000. It was made by sampling, cropping, and resizing the CelebA dataset images and augmenting it with textual annotations. However, this approach had two significant issues. Firstly, the number of facial features covered in the descriptions was again limited to \textbf{3-5 facial features per annotation of an image} that is insufficient to learn various facial features. Secondly, the cropping removed some of the essential external facial features like hair color, head shape, and accessories, and these were consequently absent in the descriptions given by the annotators.

\section{Dataset}

\begin{table*}[!t]
    \caption{Crowdsourced facial features part of the MAST dataset. The "Values" row indicates the possible value(s) considered corresponding to that feature.}
    \label{table:attributes}
    \begin{center}
        \begin{small}
            \begin{tabular}{M{2cm}M{1.75cm}M{1.75cm}M{1.75cm}M{1.75cm}M{1.75cm}M{1.75cm}M{1.75cm}M{1.75cm}}
                \toprule
                \toprule
                \textsc{\textbf{Attribute}} & \textbf{Face Shape} &
                \textbf{Eyebrows Size} & \textbf{Eyebrows Shape} & \textbf{Eyes Color} & \textbf{Eyes Size} & \textbf{Eyes Shape} & \textbf{Skin Complexion} \\
                \midrule
                \midrule
                \textsc{\textbf{Values}} & round, oval, square, rectangle, triangle & thin, normal, thick & straight, round, angled & black, brown, blue, green, grey & small, normal, big & almond, deep-set, upturned, downturned & white, fair, pale, wheatish, brown, black\\
                \bottomrule
            \end{tabular}
        \end{small}
    \end{center}
    \vspace{-0.48cm}
\end{table*}

We use the crowdsourcing platform, the CelebA dataset, the CelebA-HQ dataset, and the Microsoft Face API for the dataset creation, comprising face images and structured text descriptions.

\paragraph{CelebA-HQ dataset and Microsoft Face API} The CelebA dataset consists of about 200K celebrity images, each with 40 binary attribute annotations about facial features. The CelebA-HQ dataset is a high-quality version of the CelebA dataset, consisting of 30,000 images of up to 1024x1024 resolution without the binary attributes. Subsequently, we map the two datasets to get the face attributes for these high-resolution images. The Microsoft Face API detects faces in an image. It then returns a JSON object containing its various attributes like hair color, gender, facial hair, glasses, emotion, occlusion, and accessories with an accuracy of 90-95\%. 

\paragraph{Crowdsourcing Labeling Platform} Before coming up with our platform to annotate images, we explored several existing crowdsource labeling platforms. To satisfy our needs of a user-friendly, time and cost-efficient, compatible, and fully controlled annotating platform, we developed our platform using MEAN stack (MongoDB, Express, Angular, and Nodejs) and hosted it online. It collects seven facial features: face shape, eye size, eye shape, eye color, eyebrow size, eyebrow shape, and skin color. Our platform provides information about the task, motivation behind it, and steps to label an image. To ensure anonymity and privacy of annotators, it allows them to start labeling without signing up or signing in. Once an annotator starts the labeling session, a 10-minute timer is set on the backend, preventing the same image from reappearing. At the end of the timer, if the image is not labeled, it is brought back for annotation. Every image is labeled only once by an annotator. We employ radio buttons with attribute values for each facial attribute instead of a text box to make the labeling process faster. These attribute values are based on a prior survey campaign we had conducted to determine the most relevant attribute values from people's perspectives. This significantly streamlined the process of labeling. Finally, we provide a feedback page at the end of labeling to gather suggestions and continuously improve the platform. The platform is accessible from the following link \href{http://face-descriptions.herokuapp.com/}{\color{blue}http://face-descriptions.herokuapp.com/}

\paragraph{MAST dataset} We consolidate a new dataset consolidation and name it Multi-Attributed and Structured Text-to-face (MAST) dataset. MAST dataset is hosted on Zenodo and has this \href{https://doi.org/10.5281/zenodo.3865238}{\color{blue}Digital Object Identifier}. The motivation is to have a large corpus of high-quality face images with fine-grained and attribute-focused textual annotations. For the MAST dataset, we manually select a subset of 1993 images from the CelebA-HQ dataset. We ensure that each image has a single face without any occlusions or filters. Then, we collect facial attributes from the three sources, namely our crowdsourcing platform, the CelebA-HQ dataset, and the Microsoft Face API. The dataset consists of 30 attributes, out of which, we crowdsource 7 attributes and take 23 attributes from the CelebA-HQ dataset and Microsoft Face API. Every facial description uses 15 (7 crowdsourced + 8 or more for the other two sources) facial attributes on an average to describe the face. The remaining attributes describe facial hair, makeup, facial occlusions, and accessories that may or may not be present in every face description. The purpose of crowdsourcing is two-fold. Firstly, it provides a means to experimentally verify our hypothesis that an increase in the number of attributes in annotations leads to better text-to-face synthesis. Secondly, it is a means to procure detailed annotations for certain important facial features like face shape, eyes, eyebrows, and skin color missing in the latter two sources. These features are prominent to the naked eye and introduce a myriad of variations in the generated faces. Table~\ref{table:attributes} shows the crowdsourced attributes and their corresponding values.

At a lower level of abstraction, every facial feature's value can be considered a binary attribute. For example, blue eyes and black eyes are two different binary attributes in textual form. Accordingly, the 7 crowdsourced attributes comprise 29 binary attributes, and the remaining 23 attributes comprise 42 binary attributes. Further, on calculating the number of values per facial attribute, we observe 4 values on average for every crowdsourced attribute. In contrast, each of the remaining attributes has 1 value on average. Thus the crowdsourced attributes are more detailed.

\paragraph{Baseline dataset} We select 1993 more images from the CelebA-HQ dataset and name this subset as the Baseline dataset. This dataset only consists of the 23 attributes collected from the CelebA-HQ dataset and the Microsoft Face API. Every face description in the baseline dataset uses 8 facial attributes, on an average, to describe the face. As the baseline dataset is used as a basis of reference to set the benchmark FID, FSS, and FSD scores on the MAST dataset, we ensure that the attribute-wise sample distribution for both the MAST and baseline dataset is nearly the same for the 23 common attributes. 

\begin{figure}[!b]
\vspace{-0.4cm}
    \centering
    \includegraphics[width=0.85\linewidth, height=5.4cm]{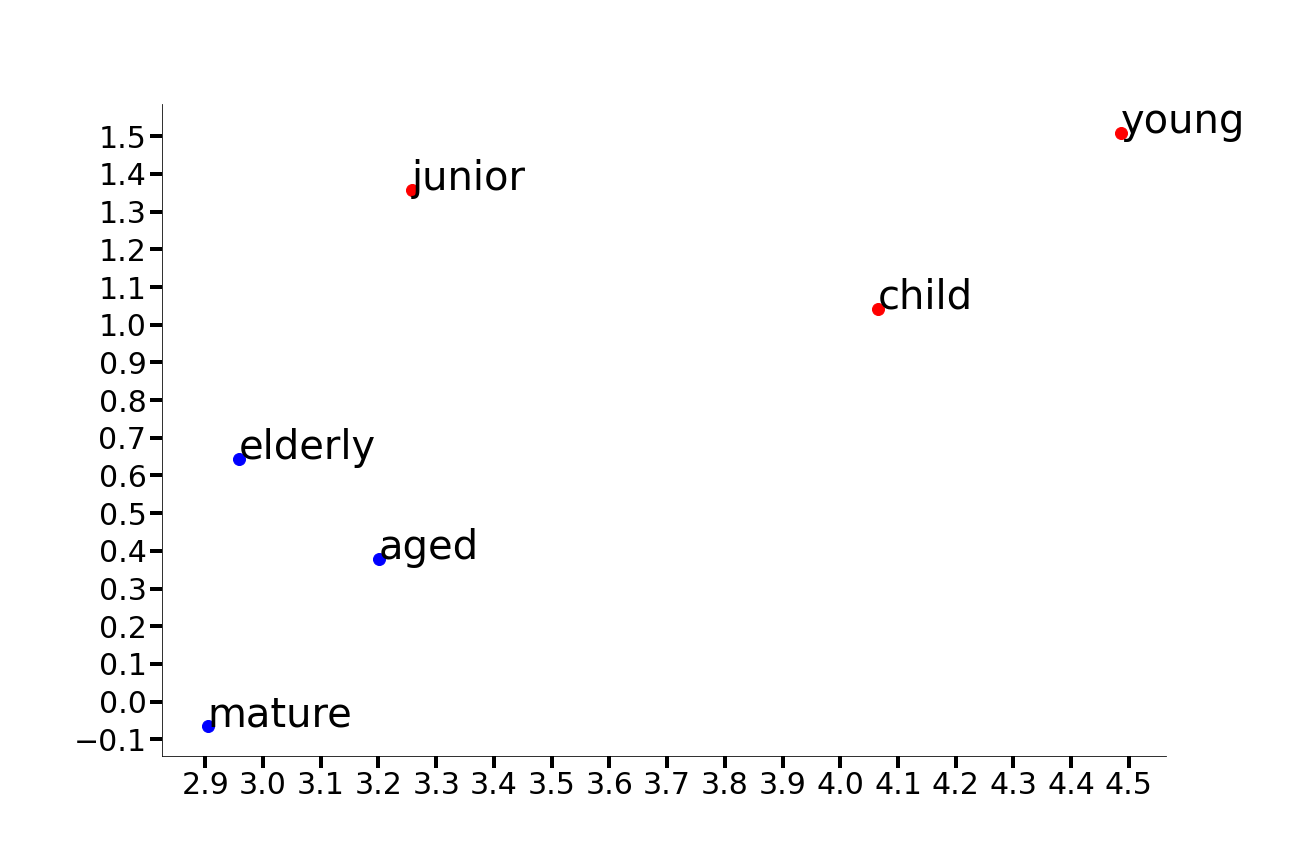}
    \caption{Visualizing cluster of some age-related words' GloVe embeddings in 2D space using Principle Component Analysis}
    \label{fig:glove-vector}
    \vspace{-0.4cm}
\end{figure}

\begin{figure*}[!t]
    \centering
    \includegraphics[width=0.95\linewidth,height=7cm]{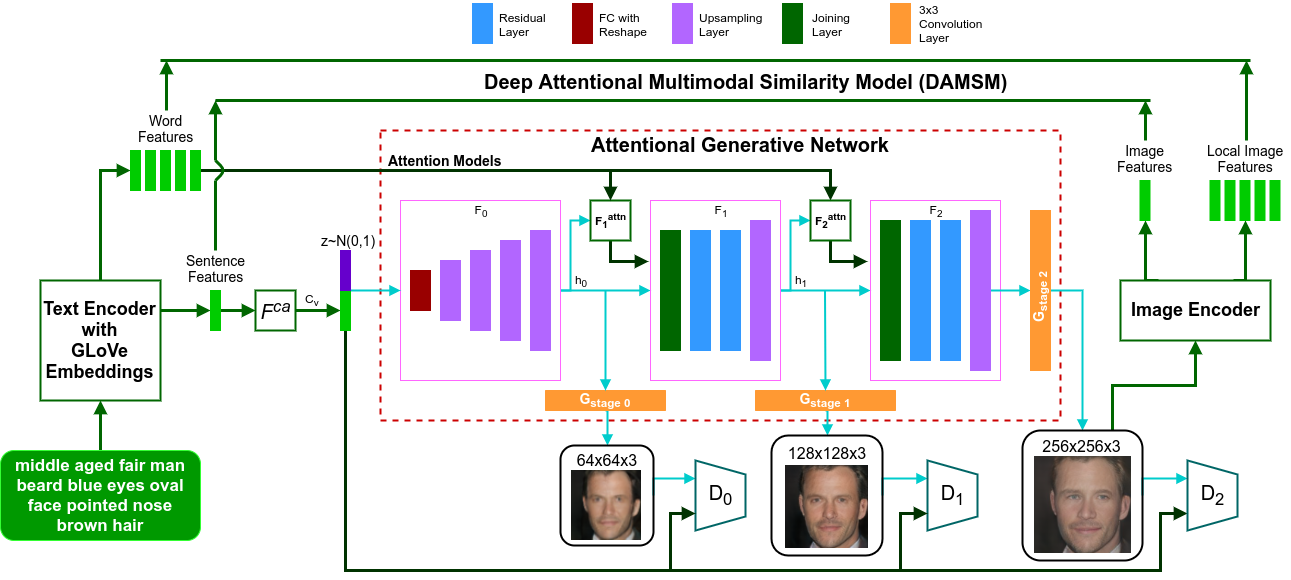}
    \caption{Architecture of the AttnGAN used. Initially, we train the text and image encoders, where GLoVe embeddings help extract semantic word vectors. Later, we train the three-stage attentional generative network using the multi-modal context vectors generated from those encoders.}
    \label{fig:architecture}
\end{figure*}

\paragraph{Dataset Processing} After consolidating the MAST and baseline dataset, we process them to make them suitable for training. Fig.~\ref{fig:sample-dataset} shows some of the face descriptions along with their face images. We create 5 textual descriptions for each face and feed them as input to the AttnGAN model. Each description consists of a random concatenation of the facial attributes to approximate the free-flowing textual descriptions. In textual description punctuation, and stopwords do not describe a face and can be neglected. Further, the random concatenation ensures that the attributes appear in a random order as they appear in a regular description. For example, the following three free-flowing descriptions are for the same face  \\
\\
\textbf{Description-1:} An old man with gray hair and blue eyes. He is smiling. \\
\textbf{Description-2:} A smiling old man. He has blue eyes and gray hair. \\
\textbf{Description-3:} A blue-eyed man. He is smiling. He is an old man with gray hair. \\
\\
In these three descriptions, \textit{$<$old$>$, $<$man$>$, $<$gray hair$>$, $<$blue eyes$>$, $<$smiling$>$} are the descriptors. After text processing, we get \\
\\
\textbf{Processed Description-1:} old man gray hair blue eyes smiling \\
\textbf{Processed Description-2:} smiling old man blue eyes gray hair \\
\textbf{Processed Description-3:} blue-eyed man smiling old man gray hair \\
\\
The above three processed descriptions are similar to the textual descriptions in the MAST dataset, as shown in Fig.~\ref{fig:sample-dataset}. Furthermore, we use the textual form of attributes to avoid the constraint of a fixed corpus of face attributes. In the case of the attribute-editing approach, descriptive words outside the fixed corpus are not considered. However, the use of textual attributes to form sentences enables the model to learn relationships between attributes. The GloVe embeddings help us sample unseen attribute values from the word cluster in the embedding space. E.g., a model trained on age attribute values like "elderly" and "aged" will be able to understand a new similar value like "mature" as these values form a close cluster in the GloVe vector space as shown in Fig.~\ref{fig:glove-vector}.

\section{AttnGAN Model}
The architecture of the AttnGAN, as shown in Fig~\ref{fig:architecture} consists of the following two components.

\begin{figure}[!t]
    \centering
    \includegraphics[width=0.75\linewidth]{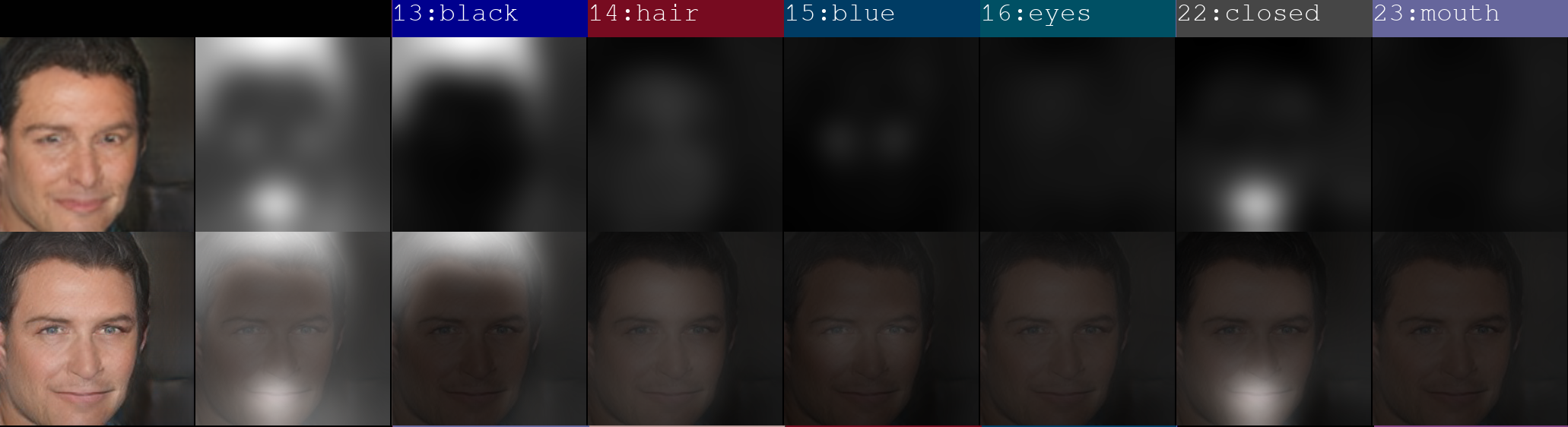}
    \caption{Attention maps of a generated face image after training on MAST dataset. Above image is of 128x128 resolution while the lower images is of 256x256 resolution.}
    \label{fig:attention-maps}
    \vspace{-0.55cm}
\end{figure}

\paragraph{Text and Image Encoder} 
Deep Attentional Multimodal Similarity Model(DAMSM) \cite{xu2018attngan} is used to train a text-encoder and image-decoder framework jointly. It maps the word embeddings for each word in the input sentence to each sub-region vectors in the given image and correlates the sentence embedding of the entire text to the feature vector of the whole image. This is shown in Fig.~\ref{fig:attention-maps}. Contrary to the pre-trained text-encoder used in the original AttnGAN, we use GLoVe vectors in the embedding layer. These vectors cover a large vocabulary and render semantic relationships among the words, suitable for capturing numerous complex facial features. Besides, this also allows the model to relate new words, which are not present in the training dataset, with attribute values. Then, the extracted text embeddings of the face descriptions and the feature vectors of the face images, respectively, are used by the next component.

\paragraph{Generator and Discriminator} 
The entire GAN model uses 3 Generators and 3 Discriminators that work on 64x64, 128x128, and 256x256 resolution images, as shown in Fig.~\ref{fig:architecture}. The initial generators help construct dominant facial features like head shape, skin color, and hairstyle. The subsequent ones construct fine-grained facial features like eyes, color, expression, and eyebrows shape, as shown in Fig.~\ref{fig:attention-maps}. 

\section{Metrics}
For the quantitative evaluation of the generated faces, we use the following four metrics - FID, FSD, and FSS. FID evaluates the quality of the generated faces. FSD and FSS metrics measure the similarity of the generated face with the real image.

\paragraph{FID} This metric calculates the distance between the feature vectors of the real and generated samples. FID metric extracts 2048-D feature vectors using the final coding layer of the Inception-net-v3 trained on the ImageNet dataset \cite{deng2009imagenet}. FID is calculated using 
\begin{equation}
    \resizebox{0.7\linewidth}{!}{
        $d^2((\textbf{m},\textbf{C}),(\textbf{m}\textsubscript{w},\textbf{C}\textsubscript{w})) = ||\textbf{m}-\textbf{m}\textsubscript{w}||^2_2 + Tr(\textbf{C} + \textbf{C}\textsubscript{w} - 2(\textbf{CC}\textsubscript{w})^\frac{1}{2})$
    }
\end{equation}
where \textit{G(m,C)} and \textit{G(m\textsubscript{w},C\textsubscript{w})} are Gaussians representing real and generated image distributions, \textbf{m} and \textbf{m}\textsubscript{w} are feature-wise mean of the real and generated images, and \textbf{C} and \textbf{C}\textsubscript{w} are covariance matrix of real and generated distribution respectively. A lower FID score represents better quality of generated images. Conversely, a higher score represents lower quality images. 

\paragraph{FSD and FSS}
FSD and FSS metrics use the Inception-net-v1 trained on the VGGFACE2 dataset \cite{cao2018vggface2} to extract feature vectors of real and generated faces. FSD is calculated using Equation~\ref{eq:fsd} and FSS using Equation~\ref{eq:fss}
\begin{equation}
    FSD = \frac{1}{N}\sum_{i=0}^{N}|Facenet(F\textsubscript{\textit{Gi}})-Facenet(F\textsubscript{\textit{G}}T\textsubscript{\textit{i}})|
    \label{eq:fsd}
\end{equation}
\begin{equation}
    FSS = \frac{1}{N}\sum_{i=0}^{N}cos(Facenet(F\textsubscript{\textit{Gi}})-Facenet(F\textsubscript{\textit{G}}T\textsubscript{\textit{i}}))
    \label{eq:fss}
\end{equation}

Here \textit{Facenet()} means using the aforementioned Inception-net-v1 model to extract a semantic vector of the input face, \textit{F\textsubscript{Gi}} means one of the generated faces, \textit{F\textsubscript{G}T\textsubscript{i}} means the ground-truth of the synthesized face image. A lower FSD score and a higher FSS percentage imply that the generated faces are similar to the ground truth faces. In other words, the generated faces accurately capture the facial features present in the real faces.

\section{Experimental Results}
\subsection{Training}
We train two GAN models based on the AttnGAN architecture. The first model is trained on the MAST dataset, and the second one is trained on the baseline dataset. Before training the models, the datasets are split into two sets - a training set consisting of 1793 images and a test set consisting of 200 images. During training, we observe that after a certain number of epochs, the generated faces' quality decreases. This is due to the gradient diminishing problem because the discriminator is learning too well and faster than the generator. To solve this problem, we use the One-sided label smoothing technique by replacing the positive class labels 1 with 0.9 in the discriminator model \cite{salimans2016improved}. We also train the discriminator on alternative epochs to stabilize the GAN training.

\subsection{Experimental investigations on the MAST dataset}

\paragraph{Evaluation details} Firstly, we use the model trained on the MAST dataset to generate ten batches of images of the corresponding test set, where a batch consists of all the images in the test set. Then, we determine the value for these metrics on each batch and finally report the average value of all the batches. Similarly, we evaluate the baseline dataset using the model trained on it. We ensure that all other experimental parameters, except the number of attributes, are the same for both datasets. 

\begin{figure}[!t]
\begin{center}
\centerline{\includegraphics[width=7.6cm, height=9.6cm]{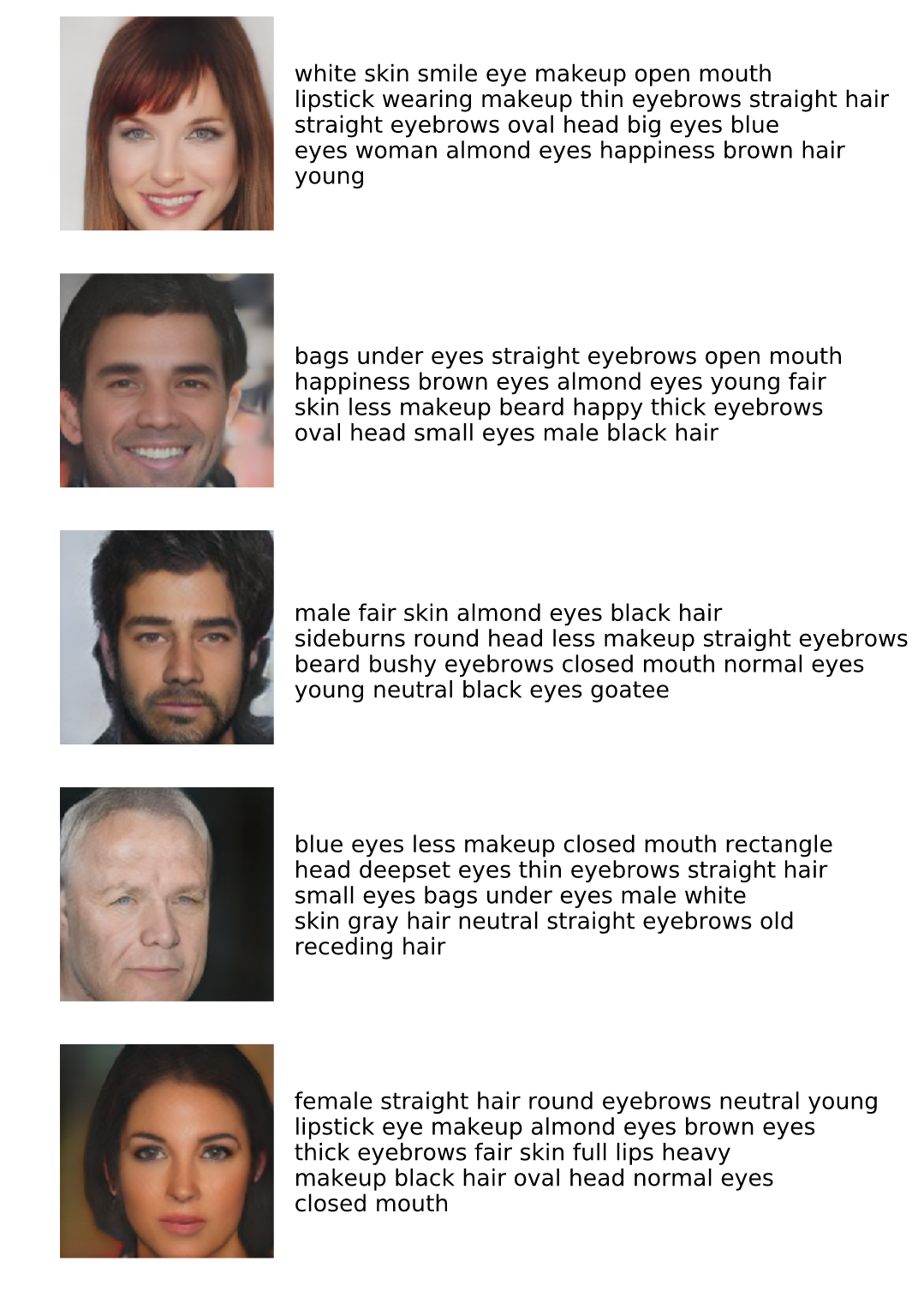}}\par
\caption{Some Generated Face images along with their textual descriptions.}
\label{fig:sample-dataset-1}
\end{center}
\vspace{-0.92cm}
\end{figure}
\begin{figure*}[!t]
	    \begin{minipage}{.45\columnwidth}
		\includegraphics[width=7.6cm, height=9.6cm]{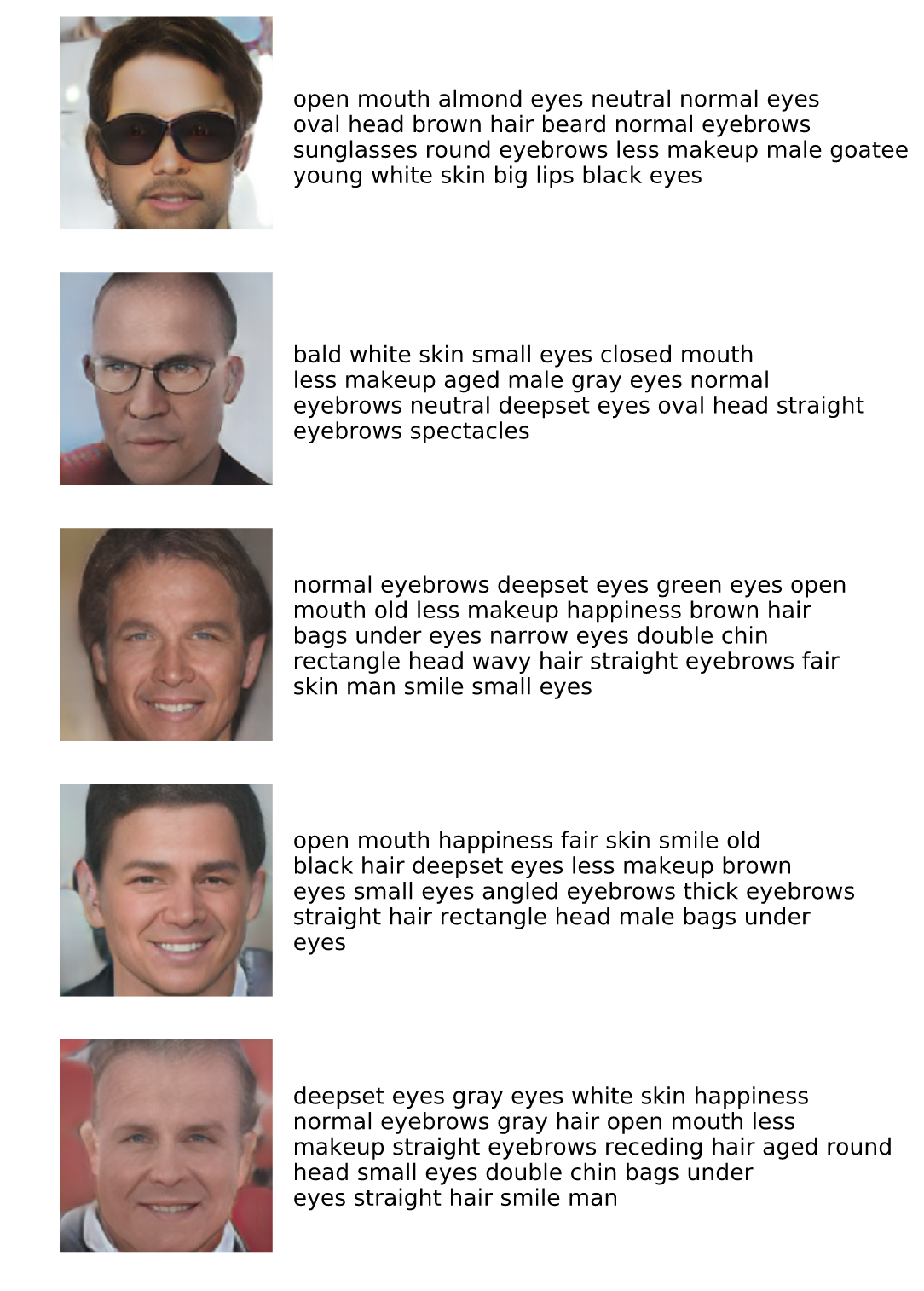}
	\end{minipage}
	\hspace{5cm}
	\begin{minipage}{.45\columnwidth}
		\includegraphics[width=7.6cm, height=9.6cm]{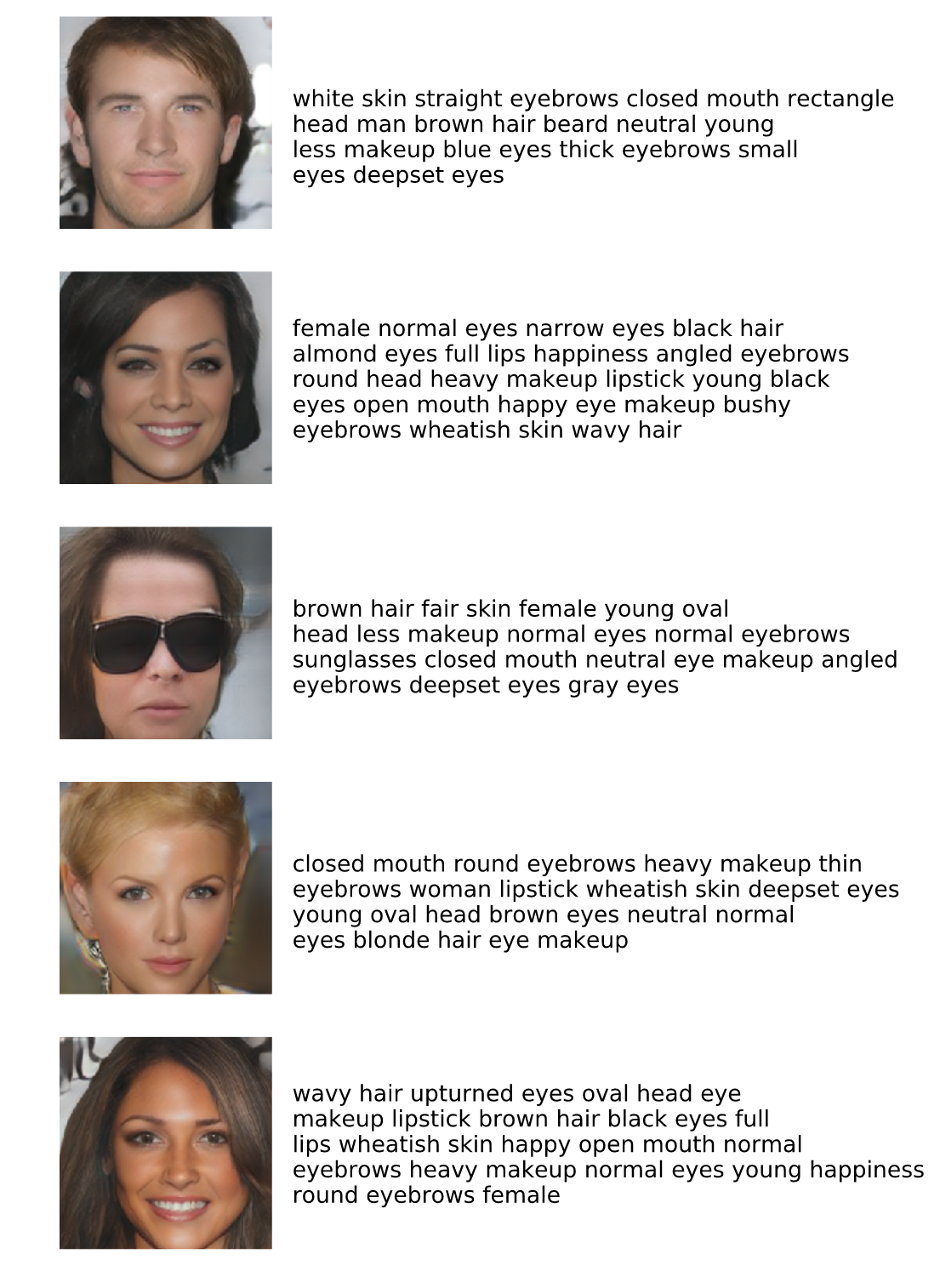}
	
	\end{minipage}%
	\caption{Some Generated Face images along with their textual descriptions.}
    \label{fig:sample-dataset-2}
	\vspace{-0.51cm}
\end{figure*}

\paragraph{Qualitative Analysis of Generated Images} The generated images are realistic, fine-grained, and encompass multiple facial features, as shown in Fig.~\ref{fig:sample-dataset-1} and Fig.~\ref{fig:sample-dataset-2}. These photos are samples generated from the MAST test dataset and show variations in gender, age, expression, skin complexion, hair color, eye color, and glasses (spectacles or sun-glasses).


\begin{table}[!t]
    \caption{Comparison of the MAST dataset with the baseline dataset to set baseline metric scores for the MAST dataset}
    \label{table:comparision-valid}
    \begin{center}
        \begin{small}
            \resizebox{\linewidth}{!}{%
                \begin{tabular}{M{1.4cm}cccccc}
                    \toprule
                    \textsc{\textbf{Network}} & \textsc{\textbf{Dataset}} & \textsc{\textbf{Attributes}} &
                    \textsc{\textbf{FID}} & \textsc{\textbf{FSD}} & \textsc{\textbf{FSS(\%)}} \\
                    \midrule
                    AttnGAN & Baseline & 23 & 67.84 & 1.095 & 51.18 \\
                    \midrule
                    AttnGAN & MAST & \textbf{30} & \textbf{54.09} & \textbf{1.080} & \textbf{60.42} \\
                    \bottomrule
                \end{tabular}
            }
        \end{small}
    \end{center}
\end{table}

\begin{table}[!t]
    \caption{Comparison of text-to-face datasets}
    \label{table:comparision}
    \vskip 0.10in
    \begin{center}
        \begin{small}
            \resizebox{\linewidth}{!}{%
                \begin{tabular}{ccM{1.4cm}M{1.4cm}cccc}
                    \toprule
                    \textbf{Network} & \textbf{Dataset} & \textbf{Training Samples} & \textbf{Testing Samples} & \textbf{FID} & \textbf{FSD} & \textbf{FSS(\%)} \\
                    \midrule
                    AttnGAN & SCU-Text2Face & 1000 & 200 & \textbf{45.56} & 1.269 & 59.28 \\
                    \midrule
                    AttnGAN & MAST & 1793 & 200 & 54.09  & \textbf{1.080} & \textbf{60.42} \\
                    \bottomrule
                \end{tabular}
            }
        \end{small}
    \end{center}
    \vspace{-0.5cm}
\end{table}

\paragraph{Quantitative Analysis of Generated Images} We compare the metric scores obtained on the MAST dataset and the baseline dataset, shown in Table~\ref{table:comparision-valid}. Under the same experimental setup, the MAST dataset outperforms the baseline dataset across all the metrics. The difference between the two datasets is 7 crowdsourced attributes that present only in the MAST dataset. These results indicate that the face generation from textual description can be improved by increasing facial attributes.
\paragraph{Comparison with other datasets} To our knowledge, the only comparable text-to-face dataset, which has reported FID, FSD, and FSS scores, is the SCU-Text2Face dataset. There are two major differences between the MAST and SCU-Text2Face dataset. Firstly, in the MAST dataset, we consider the face structure, external features like hair and accessories, and more detailed internal features than only an internal face crop used in the SCU-Text2Face. Secondly, each annotation in the MAST dataset uses 15 or more attributes, whereas there are only 3-5 attributes in each annotation of the SCU-Text2Face. 
As we did not have access to the SCU-Text2Face dataset, we used the published results for comparison, keeping other parameters like the test set size of 200 images and the GAN model. The comparison is shown in Table~\ref{table:comparision}. As a higher FSS score is desirable, the MAST dataset improves the FSS score on the SCU-Text2Face dataset by \textbf{1.92\%}. On the other hand, as a lower FSD score is desirable, the MAST dataset decreases the FSD score on the SCU-Text2Face dataset by \textbf{14.89\%} respectively. However, FID scores reported for SCU-Text2Face are better. 

We experimentally investigate this discrepancy in the FID scores in the subsequent section.

\begin{table}[!b]
    \caption{Evaluating FID, FSD and FSS metrics on the MAST and baseline dataset for different sample size.}
    \label{table:metrics}
    \begin{center}
        \begin{small}
            \begin{tabular}{M{0.88cm}M{0.82cm}M{0.82cm}M{0.82cm}M{0.82cm}M{0.82cm}M{0.82cm}M{0.82cm}M{0.82cm}}
                \toprule
                \textbf{Sample Size} & \multicolumn{2}{c}{\textbf{FID}} &  \multicolumn{2}{c}{\textbf{FSD}} & \multicolumn{2}{c}{\textbf{FSS(\%)}} \\
                \textbf{(Images)} &
                \textbf{MAST} &
                \textbf{Baseline} &
                \textbf{MAST} &
                \textbf{Baseline} &
                \textbf{MAST} &
                \textbf{Baseline} \\
                \midrule
                300 & 49.059 & 61.52 & 1.003 & 1.100 & 64.363 & 52.23 \\
                \midrule
                600 & 39.647 & 51.97 & 0.980 & 1.053 & 64.620 & 52.78 \\
                \midrule
                900 & 34.775 & 48.01 & 0.995 & 1.053 & 63.100 & 52.42 \\
                \midrule
                1000 & 33.593 & 47.20 & 0.999 & 1.061 & 63.615 & 52.34 \\
                \midrule
                1200 & 31.675 & 46.52 & 0.996 & 1.080 & 63.853 & 51.28 \\
                \midrule
                1500 & 29.486 & 45.14 & 1.005 & 1.075 & 63.704 & 50.13 \\
                \midrule
                1793 & 27.813 & 43.70 & 1.011 & 1.082 & 63.317 & 49.67 \\
                \bottomrule
            \end{tabular}
        \end{small}
    \end{center}
\end{table}

\subsection{Effect of Dataset size on FID}
\label{section:quality-metric}
\paragraph{Evaluation details} To test the sensitivity of FID on the dataset size, we need to evaluate these scores at different sample sizes for which we need a large dataset. Therefore, we use the training with 1793 samples rather than the test set with 200 samples.

\begin{figure}[!t] 
    \centering
\includegraphics[width=0.63\linewidth, height=4cm]{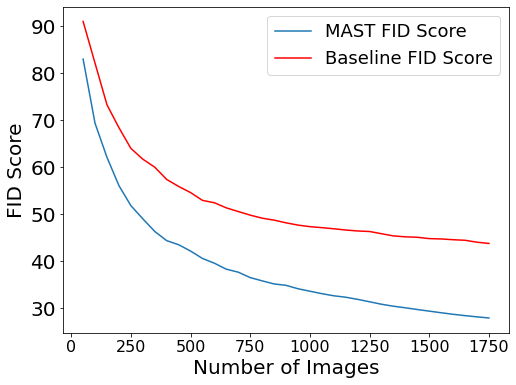} 
  \caption{Evaluation of FID  on the MAST Train and Baseline Train dataset.}
  \label{fig:train-valid-graphs} 
  \vspace{-0.47cm}
\end{figure}

We take the model trained on the MAST dataset and evaluate these scores on a subset of training samples - 300, 600, 900, 1000, 1200, 1500, and 1793. Each subset is statistically similar to the MAST training set. For evaluation, we generate ten batches of images for each subset. Then, we determine the value for these metrics on each batch and finally report the average value of all the batches. A similar evaluation is performed using the baseline dataset and the corresponding model trained on it, as shown in Table~\ref{table:metrics}

We use inductive reasoning to prove that FID is sensitive to dataset size. The premise is that when a metric is evaluated on statistically similar but different size subsets of the entire training dataset, the scores achieved must remain nearly constant. If the premise is false, then the metric is sensitive to dataset size. In the context of text-to-face synthesis, we use FSD, FSS, and FID metrics for conducting the inductive reasoning. 

\paragraph{Results} On evaluating the percentage standard deviation for FSD and FSS scores shown in Table~\ref{table:metrics}, FSD and FSS scores on MAST datasets have 0.91\% and 0.78\% deviation about the mean, and FSD and  FSS scores on baseline dataset have 0.65\% and 1.12\% deviation about the mean. By allowing 1-2\% variation in scores about the mean, these scores support the premise that metric scores evaluated on statistically similar but different size subsets of the training set must remain nearly constant.

However, on performing the same analysis for the FID metric, we obtain 19.06\% and 15.57\% about the mean for the MAST dataset and the baseline dataset, respectively. This deviation is much higher than the acceptable error limit of 1-2\%. Moreover, FID decreases by 43.30\% from the initial evaluation at a sample size of 300 to the final evaluation on the entire MAST training set. We observe a similar trend on the baseline dataset. Thus, FID scores decrease with an increase in dataset size, \emph{irrespective of the dataset being used}, as shown in the graphs shown in Fig.~\ref{fig:train-valid-graphs}. As lower FID scores are preferable, this metric suffers from under-evaluation at smaller dataset sizes. Therefore, this is the reason for the discrepancy in FID scores shown in Table~\ref{table:comparision}.
Even though the FID scores reported in Table~\ref{table:comparision-valid} are underestimated, by extrapolating the curve in graph \ref{fig:sample-dataset}, we can safely claim that evaluation on a larger test set will result in better scores. Hence, the reported FID score can serve as a benchmark FID score for the MAST dataset.

\section{Conclusion and Future Work}
This paper empirically proves that text-to-face synthesis using GANs can be improved qualitatively and quantitatively by increasing the number of attributes in each facial description. We demonstrated that a text-to-face synthesis GAN trained on structured and attribute rich textual learns several facial features, as shown in Fig.~\ref{fig:sample-dataset-1} and Fig.~\ref{fig:sample-dataset-2}. We present a new data consolidation Multi-Attributed and Structured Text-to-face (MAST) dataset and make it available to researchers to experiment and build upon. 
Finally, we report the benchmark scores of \textbf{54.09 FID, 1.080 FSD, and 60.42\% FSS} achieved by training an AttnGAN model on the MAST dataset.

Due to the limited availability of computing resources, we restricted the generative adversarial network used in this paper to synthesize images of size 256x256x3. However, the annotations made available in the MAST dataset can be mapped to the 1024x1024x3 images provided in the CELEB-A HQ dataset.  Further, the FID metric measures the synthesized image's quality by comparing its distribution to the real images' distribution. However, this measurement does not directly capture the effectiveness and reasonableness of the generated output mode from the conditioned input mode, in this case, a face image from the text. As part of our ongoing work, we are experimenting with a new metric that can directly evaluate the quality of faces generated from the text as a better alternative to FID. We are also experimenting with other GAN architectures for text-to-face synthesis. Finally, we are also continuously increasing the size and the number of attributes in the MAST dataset through our crowdsourcing platform to further improve the face generation of GANs and encourage the computer vision community to use it for their tasks. 

\bibliography{paper.bib}
\bibliographystyle{IEEEtran}

\end{document}